\documentclass{article}
\usepackage{iclr2027_conference,times}
\usepackage{hyperref}
\usepackage{url}

\definecolor{linkcolor}{HTML}{307EF1}
\hypersetup{colorlinks=true, linkcolor=linkcolor, citecolor=linkcolor,
            urlcolor=linkcolor, filecolor=linkcolor}
            
\usepackage{graphicx}
\usepackage{amsmath,amssymb}
\usepackage{booktabs,multirow,environ}
\usepackage[table]{xcolor}
\usepackage[most]{tcolorbox}
\newcommand{\figref}[1]{Figure~\ref{#1}}
\newcommand{\tabref}[1]{Table~\ref{#1}}

\newcommand{\secrefc}[1]{Section~\ref{#1}}
\newcommand{\apprefc}[1]{Appendix~\ref{#1}}
\tcbset{promptbox/.style={colback=gray!7,colframe=gray!35,
  fontupper=\small\raggedright,boxrule=0.35pt,arc=2pt,
  left=7pt,right=7pt,top=5pt,bottom=5pt,breakable}}
\newsavebox{\figfivepanelabox}
\newsavebox{\figfivelegendbox}
\newsavebox{\figfiveticksbox}
\newsavebox{\pairedleft}
\newsavebox{\pairedright}

\newcounter{tablepanel}[table]

\usepackage[TU,OT1]{fontenc}
\newcommand{\tablefont}{\normalfont}
\newcommand{\gain}[1]{\cellcolor{green!11}#1}
\newcommand{\drop}[1]{\cellcolor{red!7}#1}
\newcommand{\tablebody}{%
  \arrayrulecolor{black}\setlength{\arrayrulewidth}{0.3pt}\tablefont\fontsize{9}{10.8}\selectfont%
  \renewcommand{\arraystretch}{1.06}%
  \setlength{\tabcolsep}{3pt}%
  \setlength{\aboverulesep}{1.2pt}%
  \setlength{\belowrulesep}{1.2pt}%
  \setlength{\heavyrulewidth}{0.65pt}%
  \setlength{\lightrulewidth}{0.4pt}%
}
\newcommand{\compacttablebody}{\tablebody\fontsize{8.5}{10.2}\selectfont\setlength{\tabcolsep}{1.5pt}\renewcommand{\deltasize}{\fontsize{6.5}{8}\selectfont}}

\newsavebox{\tablemeasurebox}
\newcount\tablecolumngaps
\NewEnviron{fittedtabular}[3]{%
  \begingroup%
  \setlength{\tabcolsep}{0pt}%
  \sbox{\tablemeasurebox}{\begin{tabular}{#3}\BODY\end{tabular}}%
  \tablecolumngaps=\numexpr2*#2\relax%
  \setlength{\tabcolsep}{\dimexpr(#1-\wd\tablemeasurebox)/\tablecolumngaps\relax}%
  \begin{tabular}{#3}\BODY\end{tabular}%
  \endgroup%
}
\definecolor{tablegain}{HTML}{21804A}
\definecolor{tabledrop}{HTML}{B64040}
\newcommand{\deltasize}{\fontsize{7.5}{9}\selectfont}

\definecolor{geminiblue}{HTML}{3186FF}
\definecolor{geminigreen}{HTML}{00B95C}
\definecolor{geminiyellow}{HTML}{FBBC04}
\definecolor{geminired}{HTML}{FC413D}
\definecolor{qwenbrand}{HTML}{5945D0}
\definecolor{glmbaseline}{HTML}{E2E4E8}
\definecolor{bagelbrand}{HTML}{F59E0B}
\colorlet{qwenbaseline}{qwenbrand!30!white}
\colorlet{bagelbaseline}{bagelbrand!30!white}
\colorlet{geminibluefill}{geminiblue!30!white}
\colorlet{geminigreenfill}{geminigreen!30!white}
\colorlet{geminiyellowfill}{geminiyellow!30!white}
\colorlet{geminiredfill}{geminired!30!white}
\pgfdeclarehorizontalshading{geminirow}{10pt}{%
  color(0pt)=(geminibluefill);
  color(33pt)=(geminigreenfill);
  color(67pt)=(geminiyellowfill);
  color(100pt)=(geminiredfill)}
\makeatletter
\newcommand{\geminibackground}{%
  \makebox[0pt][l]{\hspace*{-\tabcolsep}%
    \raisebox{-\dp\@arstrutbox}[0pt][0pt]{%
      \resizebox{\linewidth}{\dimexpr\ht\@arstrutbox+\dp\@arstrutbox\relax}{%
        \pgfuseshading{geminirow}}}}}
\makeatother

\title{Can image generation help visual reasoning? An empirical study}

\author{Xueyang Yu$^{1*}$, Yifan Jing$^{1*}$, Taiming Lu$^{1}$, Linrong Cai$^{1}$, \\
\bfseries Zhipeng Cai$^{2}$, Yunyang Xiong$^{2}$, Gabriel Sarch$^{1}$, Zhuang Liu$^{1}$ \\[3.5pt]
\normalfont $^{1}$Princeton University \quad $^{2}$Meta \\
$^{*}$Equal contribution
}

\iclrfinalcopy
\begin{document}

\maketitle
\lhead{}
\renewcommand{\headrulewidth}{0pt}

 \begin{abstract}
\setlength{\parfillskip}{0pt}
Language models \textit{think in words}; then multimodal models might \textit{think in \mbox{images}}.\linebreak
We test whether a generated helper improves the answer to a visual question. A model describes the helper it needs and then answers using the original and generated images together. Across 13 vision-language model and generator pairs, generation lowers mean accuracy by 3.7 percentage points, with declines on nine of ten tasks. Two unified models, BAGEL and Janus-Pro-7B, each lose accuracy on eight tasks. On VSTAR, adding a generated image to the same planning hint lowers accuracy by 10.47 points. Repeated image edits yield different outcomes across tasks. ChartQAPro and MathVista stay below their no-generation baselines, while ERQA gains three points after ten edits.

\end{abstract}

{\small\centerline{\href{https://github.com/zlab-princeton/visual-reasoning}{\textcolor[HTML]{C71585}{\textbf{github.com/zlab-princeton/visual-reasoning}}}}}

\begin{figure}[!h]
    \vspace{0.55cm}
    \centering
    \includegraphics[width=\linewidth]{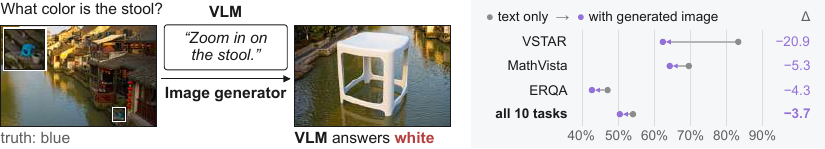}\par\nointerlineskip\vspace{.10in}
    \includegraphics[width=\linewidth]{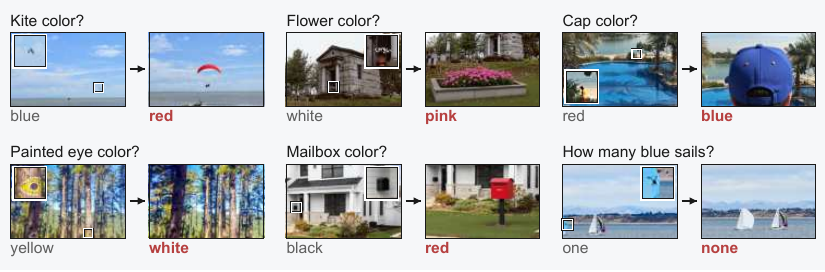}
    \vspace{-.65cm}
\caption{\textbf{Reasoning with generated images lowers accuracy.}
\textbf{Top~left:} a VLM writes a hint, an image generator draws it, and the VLM answers using the generated image, which shows a large white stool in place of the small blue one (inset).
\textbf{Top~right:} accuracy with text-only reasoning (gray) and with the generated image (purple), and the change, on three tasks and on all ten, averaged over 13 VLM--generator pairs.
\textbf{Bottom:} more failures, where the generated image changes or drops the object asked about (insets zoom in; the VLM's answers in red, the correct ones in gray).}
\label{fig:teaser}
\end{figure}

\newpage

\section{Introduction}
\label{sec:intro}

Long before we learned to write, we learned to draw: on cave walls tens of thousands of years ago, and again in every child's first crayon drawing. And we never stop. A student stuck on a geometry problem draws one extra line, and the proof falls into place. At sixteen, Einstein imagined chasing a beam of light; ten years later, that picture became special relativity. Even without paper, we rotate and rearrange images in the mind's eye, an ability we call imagination~\citep{kosslyn1996image}. Pictures are not just something we look at; they are something we think with~\citep{larkin1987diagram}. Now machines have learned to draw as well~\citep{esser2024scaling, flux2024, google2025gemini25flashimage}. If a machine can draw, can it also \emph{think} by drawing?

For now, vision-language models (VLMs) think almost entirely in words. They glance at an image once, then reason about it in long chains of text~\citep{liu2023visual, bai2025qwen2}; scaling them with chain-of-thought~\citep{shao2024visualcotadvancingmultimodal, zhang2024multimodalchainofthoughtreasoninglanguage} or long-context reinforcement learning~\citep{deepseekai2025deepseekr1incentivizingreasoningcapability, shen2025satorireinforcementlearningchainofactionthought} makes those chains longer, but no less textual. Recent visual reasoning methods let models look again, cropping, zooming, and annotating the input mid-reasoning~\citep{hu2024visual, zheng2025deepeyes, su2025openthinkimg}. But looking again is not imagining. A student who draws that extra line is not zooming into the diagram; they are adding something that was never there. Image generators can add exactly such things, and early work shows that, with curated training data in narrow settings, generated images can help~\citep{li2025imagine, wu2025govig, gu2025thinkmorph}. So we gave models a pencil and real problems. Mostly, they drew the wrong thing.

\begin{figure}[t]
    \centering
    \includegraphics[width=1\linewidth]{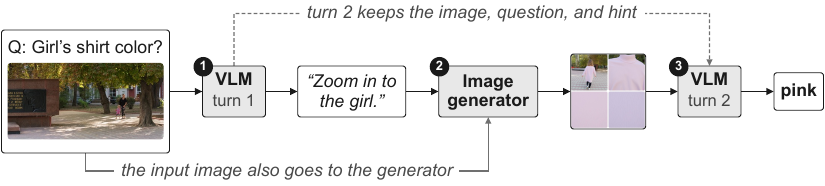}
    \vspace{-.75cm}
    \caption{
    \textbf{The Reasoning with Imagination pipeline.} The VLM writes a hint, the generator draws a helper, and the VLM answers using evidence from the original images and the helper together.}
    \label{fig:framework}

\end{figure}

Zebra-CoT~\citep{li2025zebra} supplies interleaved text and image training data, and RealUnify~\citep{shi2025realunifyunifiedmodelstruly} studies how understanding and generation support each other in unified models. We use \textit{Reasoning with Imagination} (\figref{fig:framework}) to measure how a generated helper changes the answer to a visual question. In the first turn, a VLM writes a short hint describing an image that would help, such as \emph{zoom into the top-left panel and annotate the legend}. A generator uses the hint and the input images to produce the helper. In the second turn, the VLM answers using the originals and the helper together. We compare each model with its own direct-answer baseline in both modular pipelines and unified models.

We evaluate 13 VLM--generator pipelines and two unified models, BAGEL~\citep{deng2025bagel} and Janus-Pro-7B~\citep{chen2025januspro}, on nine benchmarks spanning ten tasks. Generation lowers accuracy in 106 of 130 modular pipeline--task comparisons, with losses of up to 31.3 points on VSTAR~\citep{wu2024v}. Each unified model loses accuracy on eight tasks. BAGEL gains on MathVista~\citep{lu2024mathvista} and RoboSpatial-Compatibility~\citep{song2025robospatial}, while Janus gains on VisualPuzzles and Rover-TG.

We examine where errors enter the pipeline, hold the planning turn fixed while changing the helper, and vary the number of generation calls (\secrefc{sec:reasons}).
\begingroup
\setlength{\leftmargini}{1.2em}
\setlength{\topsep}{0pt}
\setlength{\itemsep}{2pt}
\setlength{\parsep}{2pt}
\begin{itemize}
    \item \textbf{Hint and image quality.} Of the failures we annotated, 66\% had a bad hint, 28\% a faulty image, and 6\% an error in the final inference. On VSTAR, when the hint calls for a crop, applying it to the original pixels raises accuracy by 20.93 points over a generated edit.
    \item \textbf{Test-time scaling depends on the task.} Extra text reasoning raises accuracy on two of three benchmarks. With up to ten image edits, ChartQAPro and MathVista stay below their no-generation baselines at every checkpoint. ERQA ends above baseline.
\end{itemize}
\endgroup

\vspace{5pt}

These comparisons distinguish failures in planning from failures in rendering the requested evidence. On VSTAR, even when the request stays the same, how the helper is produced changes the model's answer.

\newpage

\vspace{-8pt}
\section{Reasoning With Imagination}
\vspace{-7pt}

\textit{Reasoning with Imagination} inserts one generated image into a VLM's reasoning (\figref{fig:framework}). The input is a visual question: one or more images and a question about them. Instead of answering directly, as in the text-only baseline, the model proceeds in three steps.

\vspace{-.2cm}
\paragraph{Step 1: Hint.}
In the first turn, the VLM receives the images and the question but is instructed not to answer yet. Instead, we prompt it to write a hint of one or two sentences describing a single helper image that would make the question easier, such as ``zoom into the top-left panel and annotate the legend'' or ``rotate the first image to the right by 30 degrees.''

\vspace{-.2cm}
\paragraph{Step 2: Generation.}
An image generator draws the helper image from the hint and the input images. Generators that accept only one image edit the first input image; those that accept several see all of them. Unless stated otherwise, each question receives one helper image.

\vspace{-.2cm}
\paragraph{Step 3: Inference.}
The helper image is added to the same conversation in a second turn. The VLM then answers with the question, the original images, its hint, and the helper image all in context, and is asked to use evidence from every image. All prompts are in \apprefc{app:prompts}.

\vspace{-.2cm}
\paragraph{Modular and unified models.}
\label{sec:method:two-model}
\label{sec:method:unified}
In modular pipelines, a VLM plans and answers, and a separate generator produces the helper. In the unified setting, one model performs all three steps. For Janus-Pro-7B, we condition its generation head on the original image embeddings and the hint. The same backbone then answers from the original images, hint, and helper.

\section{Experiments}
\label{sec:experiments}

\newsavebox{\setupgridbox}
\newlength{\setupbenchwidth}
\begin{table}[!hb]
\centering
\sbox{\setupgridbox}{%
\tablebody\small
\begin{tabular}[t]{l}
\toprule
\textbf{VLM} \\
\midrule
{\renewcommand{\arraystretch}{1.413}\begin{tabular}[c]{@{}l@{}}
Qwen3-VL 4B~\citep{bai2025qwen3vl} \\
Qwen3-VL 8B~\citep{bai2025qwen3vl} \\
GLM-4.6V Flash~\citep{hong2025glm}
\end{tabular}} \\
\bottomrule
\end{tabular}%
\hspace{0.12in}\raisebox{-0.556in}{\Large$\times$}\hspace{0.12in}%
\begin{tabular}[t]{l}
\toprule
\textbf{image generator} \\
\midrule
Stable Diffusion 3.5 Large~\citep{esser2024scaling} \\
FLUX.1-Kontext Dev~\citep{flux2024} \\
Qwen-Image-Edit-2511~\citep{wu2025qwen} \\
Nano-Banana~\citep{google2025gemini25flashimage} \\
\bottomrule
\end{tabular}}%
\usebox{\setupgridbox}
\vspace{-.2cm}
\caption{\textbf{Modular pipelines.} Every VLM is paired with every image generator.}
\label{tab:models}
\vspace{.35cm}
\begingroup
\tablebody\small
\settowidth{\setupbenchwidth}{spatial and embodied}%
\setlength{\setupbenchwidth}{\dimexpr\wd\setupgridbox-\setupbenchwidth-4\tabcolsep\relax}%
\begin{tabular}{l>{\raggedright\arraybackslash}p{\setupbenchwidth}}
\toprule
\textbf{category} & \textbf{benchmarks} \\
\midrule
chart understanding & ChartQAPro~\citep{masry2025chartqapro} \\
visual search & VSTAR~\citep{wu2024v} \\
visual logic & VisuLogic~\citep{xu2025visulogic}, VisualPuzzles~\citep{song2025visualpuzzles} \\
spatial and embodied & SAT~\citep{ray2024sat}, RoboSpatial~\citep{song2025robospatial},\newline ERQA~\citep{team2025gemini} \\
STEM & MathVista-mini~\citep{lu2024mathvista} \\
cross-modal & Rover-TG~\citep{liang2025rover} \\
\bottomrule
\end{tabular}
\endgroup
\vspace{-.2cm}
\caption{\textbf{Benchmarks.} Nine benchmarks covering six kinds of visual reasoning.}
\label{tab:benchmarks}
\end{table}

\definecolor{basegray}{RGB}{246,247,249}
\definecolor{gainbg}{RGB}{232,247,235}
\definecolor{dropbg}{RGB}{252,238,238}
\renewcommand{\gain}[1]{\cellcolor{gainbg}#1}
\renewcommand{\drop}[1]{\cellcolor{dropbg}#1}

\begin{table*}[t]
\centering
\begingroup
\tablebody\normalsize
\definecolor{tablezero}{gray}{0.45}
\newcommand{\change}[2]{{\fontsize{8.5}{10}\selectfont\textcolor{#1}{#2}}}
\ifdefined\resvalw\else\newlength{\resvalw}\newlength{\reschw}\fi
\settowidth{\resvalw}{\textbf{00.0}}
\settowidth{\reschw}{\change{black}{\textminus{}00.0}}
\newcommand{\res}[2]{\makebox[\resvalw][r]{#1}\hspace{3pt}\makebox[\reschw][l]{#2}}
\newcommand{\modelindent}{\hspace{1.2em}}
\newcommand{\generatorrow}{\hspace{2.4em}}
\definecolor{geminirow}{HTML}{E4EDF8}
\pgfdeclarehorizontalshading{geminifade}{10pt}{color(0pt)=(geminirow); color(100pt)=(geminirow)}
\definecolor{qwenrow}{HTML}{ECE7F5}
\pgfdeclarehorizontalshading{qwenfade}{10pt}{color(0pt)=(qwenrow); color(100pt)=(qwenrow)}
\definecolor{glmrow}{HTML}{E5F1E8}
\pgfdeclarehorizontalshading{glmfade}{10pt}{color(0pt)=(glmrow); color(100pt)=(glmrow)}
\definecolor{bagelrow}{HTML}{F7E8EE}
\pgfdeclarehorizontalshading{bagelfade}{10pt}{color(0pt)=(bagelrow); color(100pt)=(bagelrow)}
\definecolor{janusrow}{HTML}{F9E9C0}
\pgfdeclarehorizontalshading{janusfade}{10pt}{color(0pt)=(janusrow); color(100pt)=(janusrow)}
\makeatletter
\newcommand{\modelrow}[1]{%
  \makebox[0pt][l]{\hspace*{-\tabcolsep}%
    \raisebox{-\dp\@arstrutbox}[0pt][0pt]{%
      \resizebox{\linewidth}{\dimexpr\ht\@arstrutbox+\dp\@arstrutbox\relax}{\pgfuseshading{#1}}}}}
\makeatother
\begin{fittedtabular}{\linewidth}{6}{l*{5}{c}}
\toprule
 & \makebox[0pt]{ChartQAPro} & \makebox[0pt]{VisuLogic} & \makebox[0pt]{VisualPuzzles} & \makebox[0pt]{VSTAR} & \makebox[0pt]{Rover-TG} \\
\midrule
\multicolumn{6}{l}{\textbf{VLM + generator}} \\
\modelrow{geminifade}\modelindent Gemini & \res{\textbf{66.4}}{} & \res{\textbf{36.4}}{} & \res{\textbf{46.2}}{} & \res{\textbf{88.0}}{} & \res{\textbf{41.9}}{} \\
\generatorrow + Nano-Banana & \res{58.1}{\change{tabledrop}{\textminus{}8.3}} & \res{34.1}{\change{tabledrop}{\textminus{}2.3}} & \res{34.9}{\change{tabledrop}{\textminus{}11.3}} & \res{71.7}{\change{tabledrop}{\textminus{}16.3}} & \res{22.1}{\change{tabledrop}{\textminus{}19.8}} \\
\modelrow{qwenfade}\modelindent Qwen3-VL 4B & \res{\textbf{35.6}}{} & \res{\textbf{26.3}}{} & \res{22.1}{} & \res{\textbf{81.3}}{} & \res{\textbf{31.9}}{} \\
\generatorrow + SD3.5 & \res{33.6}{\change{tabledrop}{\textminus{}2.0}} & \res{25.4}{\change{tabledrop}{\textminus{}0.9}} & \res{23.3}{\change{tablegain}{+1.2}} & \res{63.4}{\change{tabledrop}{\textminus{}17.9}} & \res{18.3}{\change{tabledrop}{\textminus{}13.6}} \\
\generatorrow + FLUX.1 & \res{35.5}{\change{tabledrop}{\textminus{}0.1}} & \res{26.0}{\change{tabledrop}{\textminus{}0.3}} & \res{22.7}{\change{tablegain}{+0.6}} & \res{77.5}{\change{tabledrop}{\textminus{}3.8}} & \res{18.3}{\change{tabledrop}{\textminus{}13.6}} \\
\generatorrow + Qwen-Image-Edit & \res{33.2}{\change{tabledrop}{\textminus{}2.4}} & \res{24.7}{\change{tabledrop}{\textminus{}1.6}} & \res{21.9}{\change{tabledrop}{\textminus{}0.2}} & \res{50.8}{\change{tabledrop}{\textminus{}30.5}} & \res{16.0}{\change{tabledrop}{\textminus{}15.9}} \\
\generatorrow + Nano-Banana & \res{35.2}{\change{tabledrop}{\textminus{}0.4}} & \res{24.9}{\change{tabledrop}{\textminus{}1.4}} & \res{\textbf{23.4}}{\change{tablegain}{+1.3}} & \res{59.7}{\change{tabledrop}{\textminus{}21.6}} & \res{21.5}{\change{tabledrop}{\textminus{}10.4}} \\
\modelrow{qwenfade}\modelindent Qwen3-VL 8B & \res{\textbf{40.8}}{} & \res{25.6}{} & \res{23.1}{} & \res{\textbf{81.6}}{} & \res{22.9}{} \\
\generatorrow + SD3.5 & \res{38.6}{\change{tabledrop}{\textminus{}2.2}} & \res{24.4}{\change{tabledrop}{\textminus{}1.2}} & \res{23.4}{\change{tablegain}{+0.3}} & \res{54.5}{\change{tabledrop}{\textminus{}27.1}} & \res{17.4}{\change{tabledrop}{\textminus{}5.5}} \\
\generatorrow + FLUX.1 & \res{38.7}{\change{tabledrop}{\textminus{}2.1}} & \res{25.1}{\change{tabledrop}{\textminus{}0.5}} & \res{22.4}{\change{tabledrop}{\textminus{}0.7}} & \res{69.6}{\change{tabledrop}{\textminus{}12.0}} & \res{18.9}{\change{tabledrop}{\textminus{}4.0}} \\
\generatorrow + Qwen-Image-Edit & \res{38.3}{\change{tabledrop}{\textminus{}2.5}} & \res{25.4}{\change{tabledrop}{\textminus{}0.2}} & \res{23.4}{\change{tablegain}{+0.3}} & \res{50.8}{\change{tabledrop}{\textminus{}30.8}} & \res{22.1}{\change{tabledrop}{\textminus{}0.8}} \\
\generatorrow + Nano-Banana & \res{40.4}{\change{tabledrop}{\textminus{}0.4}} & \res{\textbf{26.4}}{\change{tablegain}{+0.8}} & \res{\textbf{25.0}}{\change{tablegain}{+1.9}} & \res{58.3}{\change{tabledrop}{\textminus{}23.3}} & \res{\textbf{24.1}}{\change{tablegain}{+1.2}} \\
\modelrow{glmfade}\modelindent GLM-4.6V Flash & \res{\textbf{60.5}}{} & \res{\textbf{25.9}}{} & \res{\textbf{37.0}}{} & \res{\textbf{86.2}}{} & \res{\textbf{33.4}}{} \\
\generatorrow + SD3.5 & \res{56.9}{\change{tabledrop}{\textminus{}3.6}} & \res{24.8}{\change{tabledrop}{\textminus{}1.1}} & \res{34.9}{\change{tabledrop}{\textminus{}2.1}} & \res{60.7}{\change{tabledrop}{\textminus{}25.5}} & \res{28.8}{\change{tabledrop}{\textminus{}4.6}} \\
\generatorrow + FLUX.1 & \res{58.9}{\change{tabledrop}{\textminus{}1.6}} & \res{25.1}{\change{tabledrop}{\textminus{}0.8}} & \res{34.2}{\change{tabledrop}{\textminus{}2.8}} & \res{82.2}{\change{tabledrop}{\textminus{}4.0}} & \res{29.1}{\change{tabledrop}{\textminus{}4.3}} \\
\generatorrow + Qwen-Image-Edit & \res{56.7}{\change{tabledrop}{\textminus{}3.8}} & \res{25.0}{\change{tabledrop}{\textminus{}0.9}} & \res{33.9}{\change{tabledrop}{\textminus{}3.1}} & \res{54.9}{\change{tabledrop}{\textminus{}31.3}} & \res{32.1}{\change{tabledrop}{\textminus{}1.3}} \\
\generatorrow + Nano-Banana & \res{58.9}{\change{tabledrop}{\textminus{}1.6}} & \res{25.4}{\change{tabledrop}{\textminus{}0.5}} & \res{35.1}{\change{tabledrop}{\textminus{}1.9}} & \res{57.9}{\change{tabledrop}{\textminus{}28.3}} & \res{32.5}{\change{tabledrop}{\textminus{}0.9}} \\
\midrule
\multicolumn{6}{l}{\textbf{unified models}} \\
\modelrow{bagelfade}\modelindent BAGEL & \res{\textbf{29.4}}{} & \res{\textbf{41.3}}{} & \res{\textbf{26.2}}{} & \res{\textbf{67.0}}{} & \res{\textbf{25.6}}{} \\
\generatorrow + generation & \res{26.0}{\change{tabledrop}{\textminus{}3.4}} & \res{39.3}{\change{tabledrop}{\textminus{}2.0}} & \res{25.2}{\change{tabledrop}{\textminus{}1.0}} & \res{47.6}{\change{tabledrop}{\textminus{}19.4}} & \res{20.6}{\change{tabledrop}{\textminus{}5.0}} \\
\modelrow{janusfade}\modelindent Janus-Pro-7B & \res{\textbf{16.1}}{} & \res{\textbf{18.3}}{} & \res{16.1}{} & \res{\textbf{47.1}}{} & \res{10.8}{} \\
\generatorrow + generation & \res{9.8}{\change{tabledrop}{\textminus{}6.4}} & \res{10.7}{\change{tabledrop}{\textminus{}7.6}} & \res{\textbf{17.4}}{\change{tablegain}{+1.3}} & \res{42.9}{\change{tabledrop}{\textminus{}4.2}} & \res{\textbf{12.8}}{\change{tablegain}{+2.0}} \\
\bottomrule
\end{fittedtabular}%
\endgroup
\vspace{-.25cm}
\caption{
\textbf{Visual generation mostly lowers accuracy on chart, logic, visual-search, and cross-modal tasks.} Across modular VLM--generator pipelines, accuracy drops in every pipeline on ChartQAPro and VSTAR, and the few gains are small (at most 1.9 points, mostly on VisualPuzzles). BAGEL declines on all five tasks. Janus-Pro-7B gains on VisualPuzzles and Rover-TG. Bold: group-best accuracy (ties included). Small colored numbers: change from the corresponding text-only baseline in percentage points (green = gains, red = losses). Janus use unrounded accuracies.
}
\vspace{-.15cm}
\label{tab:main-1}
\end{table*}

\definecolor{basegray}{RGB}{246,247,249}
\definecolor{gainbg}{RGB}{232,247,235}
\definecolor{dropbg}{RGB}{252,238,238}
\renewcommand{\gain}[1]{\cellcolor{gainbg}#1}
\renewcommand{\drop}[1]{\cellcolor{dropbg}#1}

\begin{table*}[t]
\centering
\begingroup
\tablebody\normalsize
\definecolor{tablezero}{gray}{0.45}
\newcommand{\change}[2]{{\fontsize{8.5}{10}\selectfont\textcolor{#1}{#2}}}
\ifdefined\resvalw\else\newlength{\resvalw}\newlength{\reschw}\fi
\settowidth{\resvalw}{\textbf{00.0}}
\settowidth{\reschw}{\change{black}{\textminus{}00.0}}
\newcommand{\res}[2]{\makebox[\resvalw][r]{#1}\hspace{3pt}\makebox[\reschw][l]{#2}}
\newcommand{\modelindent}{\hspace{1.2em}}
\newcommand{\generatorrow}{\hspace{2.4em}}
\definecolor{geminirow}{HTML}{E4EDF8}
\pgfdeclarehorizontalshading{geminifade}{10pt}{color(0pt)=(geminirow); color(100pt)=(geminirow)}
\definecolor{qwenrow}{HTML}{ECE7F5}
\pgfdeclarehorizontalshading{qwenfade}{10pt}{color(0pt)=(qwenrow); color(100pt)=(qwenrow)}
\definecolor{glmrow}{HTML}{E5F1E8}
\pgfdeclarehorizontalshading{glmfade}{10pt}{color(0pt)=(glmrow); color(100pt)=(glmrow)}
\definecolor{bagelrow}{HTML}{F7E8EE}
\pgfdeclarehorizontalshading{bagelfade}{10pt}{color(0pt)=(bagelrow); color(100pt)=(bagelrow)}
\definecolor{janusrow}{HTML}{F9E9C0}
\pgfdeclarehorizontalshading{janusfade}{10pt}{color(0pt)=(janusrow); color(100pt)=(janusrow)}
\makeatletter
\newcommand{\modelrow}[1]{%
  \makebox[0pt][l]{\hspace*{-\tabcolsep}%
    \raisebox{-\dp\@arstrutbox}[0pt][0pt]{%
      \resizebox{\linewidth}{\dimexpr\ht\@arstrutbox+\dp\@arstrutbox\relax}{\pgfuseshading{#1}}}}}
\makeatother
\begin{fittedtabular}{\linewidth}{6}{l*{5}{c}}
\toprule
 & \makebox[0pt]{MathVista} & \makebox[0pt]{SAT} & \makebox[0pt]{ERQA} & \makebox[0pt]{RS-Compat.} & \makebox[0pt]{RS-Config.} \\
\midrule
\multicolumn{6}{l}{\textbf{VLM + generator}} \\
\modelrow{geminifade}\modelindent Gemini & \res{\textbf{78.6}}{} & \res{\textbf{79.3}}{} & \res{\textbf{66.8}}{} & \res{\textbf{76.2}}{} & \res{\textbf{82.9}}{} \\
\generatorrow + Nano-Banana & \res{74.5}{\change{tabledrop}{\textminus{}4.1}} & \res{76.7}{\change{tabledrop}{\textminus{}2.6}} & \res{64.3}{\change{tabledrop}{\textminus{}2.5}} & \res{\textbf{76.2}}{\change{tablezero}{0.0}} & \res{81.0}{\change{tabledrop}{\textminus{}1.9}} \\
\modelrow{qwenfade}\modelindent Qwen3-VL 4B & \res{\textbf{62.3}}{} & \res{\textbf{66.3}}{} & \res{\textbf{44.3}}{} & \res{45.7}{} & \res{\textbf{80.8}}{} \\
\generatorrow + SD3.5 & \res{53.3}{\change{tabledrop}{\textminus{}9.0}} & \res{64.7}{\change{tabledrop}{\textminus{}1.6}} & \res{41.8}{\change{tabledrop}{\textminus{}2.5}} & \res{68.6}{\change{tablegain}{+22.9}} & \res{67.5}{\change{tabledrop}{\textminus{}13.3}} \\
\generatorrow + FLUX.1 & \res{58.3}{\change{tabledrop}{\textminus{}4.0}} & \res{65.3}{\change{tabledrop}{\textminus{}1.0}} & \res{43.5}{\change{tabledrop}{\textminus{}0.8}} & \res{65.7}{\change{tablegain}{+20.0}} & \res{68.3}{\change{tabledrop}{\textminus{}12.5}} \\
\generatorrow + Qwen-Image-Edit & \res{48.8}{\change{tabledrop}{\textminus{}13.5}} & \res{60.0}{\change{tabledrop}{\textminus{}6.3}} & \res{38.5}{\change{tabledrop}{\textminus{}5.8}} & \res{\textbf{69.5}}{\change{tablegain}{+23.8}} & \res{75.6}{\change{tabledrop}{\textminus{}5.2}} \\
\generatorrow + Nano-Banana & \res{57.6}{\change{tabledrop}{\textminus{}4.7}} & \res{61.4}{\change{tabledrop}{\textminus{}4.9}} & \res{40.0}{\change{tabledrop}{\textminus{}4.3}} & \res{\textbf{69.5}}{\change{tablegain}{+23.8}} & \res{75.8}{\change{tabledrop}{\textminus{}5.0}} \\
\modelrow{qwenfade}\modelindent Qwen3-VL 8B & \res{\textbf{61.2}}{} & \res{\textbf{70.1}}{} & \res{\textbf{44.3}}{} & \res{61.6}{} & \res{80.4}{} \\
\generatorrow + SD3.5 & \res{52.4}{\change{tabledrop}{\textminus{}8.8}} & \res{64.3}{\change{tabledrop}{\textminus{}5.8}} & \res{40.0}{\change{tabledrop}{\textminus{}4.3}} & \res{\textbf{77.1}}{\change{tablegain}{+15.5}} & \res{77.2}{\change{tabledrop}{\textminus{}3.2}} \\
\generatorrow + FLUX.1 & \res{57.3}{\change{tabledrop}{\textminus{}3.9}} & \res{65.4}{\change{tabledrop}{\textminus{}4.7}} & \res{42.8}{\change{tabledrop}{\textminus{}1.5}} & \res{70.5}{\change{tablegain}{+8.9}} & \res{\textbf{81.3}}{\change{tablegain}{+0.9}} \\
\generatorrow + Qwen-Image-Edit & \res{56.9}{\change{tabledrop}{\textminus{}4.3}} & \res{61.6}{\change{tabledrop}{\textminus{}8.5}} & \res{38.5}{\change{tabledrop}{\textminus{}5.8}} & \res{\textbf{77.1}}{\change{tablegain}{+15.5}} & \res{75.6}{\change{tabledrop}{\textminus{}4.8}} \\
\generatorrow + Nano-Banana & \res{60.7}{\change{tabledrop}{\textminus{}0.5}} & \res{61.8}{\change{tabledrop}{\textminus{}8.3}} & \res{40.0}{\change{tabledrop}{\textminus{}4.3}} & \res{72.4}{\change{tablegain}{+10.8}} & \res{79.4}{\change{tabledrop}{\textminus{}1.0}} \\
\modelrow{glmfade}\modelindent GLM-4.6V Flash & \res{\textbf{83.3}}{} & \res{65.8}{} & \res{\textbf{47.8}}{} & \res{66.7}{} & \res{\textbf{79.8}}{} \\
\generatorrow + SD3.5 & \res{79.9}{\change{tabledrop}{\textminus{}3.4}} & \res{65.3}{\change{tabledrop}{\textminus{}0.5}} & \res{37.3}{\change{tabledrop}{\textminus{}10.5}} & \res{\textbf{79.1}}{\change{tablegain}{+12.4}} & \res{76.4}{\change{tabledrop}{\textminus{}3.4}} \\
\generatorrow + FLUX.1 & \res{80.1}{\change{tabledrop}{\textminus{}3.2}} & \res{64.0}{\change{tabledrop}{\textminus{}1.8}} & \res{43.0}{\change{tabledrop}{\textminus{}4.8}} & \res{78.1}{\change{tablegain}{+11.4}} & \res{79.7}{\change{tabledrop}{\textminus{}0.1}} \\
\generatorrow + Qwen-Image-Edit & \res{77.0}{\change{tabledrop}{\textminus{}6.3}} & \res{\textbf{68.0}}{\change{tablegain}{+2.2}} & \res{43.0}{\change{tabledrop}{\textminus{}4.8}} & \res{67.6}{\change{tablegain}{+0.9}} & \res{78.8}{\change{tabledrop}{\textminus{}1.0}} \\
\generatorrow + Nano-Banana & \res{80.1}{\change{tabledrop}{\textminus{}3.2}} & \res{\textbf{68.0}}{\change{tablegain}{+2.2}} & \res{43.2}{\change{tabledrop}{\textminus{}4.6}} & \res{71.4}{\change{tablegain}{+4.7}} & \res{76.4}{\change{tabledrop}{\textminus{}3.4}} \\
\midrule
\multicolumn{6}{l}{\textbf{unified models}} \\
\modelrow{bagelfade}\modelindent BAGEL & \res{41.8}{} & \res{\textbf{63.3}}{} & \res{\textbf{43.2}}{} & \res{46.7}{} & \res{\textbf{82.1}}{} \\
\generatorrow + generation & \res{\textbf{49.9}}{\change{tablegain}{+8.1}} & \res{54.7}{\change{tabledrop}{\textminus{}8.6}} & \res{40.8}{\change{tabledrop}{\textminus{}2.4}} & \res{\textbf{72.4}}{\change{tablegain}{+25.7}} & \res{73.2}{\change{tabledrop}{\textminus{}8.9}} \\
\modelrow{janusfade}\modelindent Janus-Pro-7B & \res{\textbf{37.4}}{} & \res{\textbf{46.7}}{} & \res{\textbf{34.5}}{} & \res{\textbf{34.3}}{} & \res{\textbf{71.5}}{} \\
\generatorrow + generation & \res{28.0}{\change{tabledrop}{\textminus{}9.4}} & \res{44.7}{\change{tabledrop}{\textminus{}2.0}} & \res{20.3}{\change{tabledrop}{\textminus{}14.3}} & \res{24.8}{\change{tabledrop}{\textminus{}9.5}} & \res{60.2}{\change{tabledrop}{\textminus{}11.4}} \\
\bottomrule
\end{fittedtabular}%
\endgroup
\vspace{-.25cm}
\caption{
\textbf{Modular pipelines gain most consistently on RoboSpatial-Compatibility.} Across modular VLM--generator pipelines, accuracy generally drops on MathVista, SAT, ERQA, and RoboSpatial-Configuration. BAGEL gains on MathVista and RoboSpatial-Compatibility. Janus-Pro-7B declines on all five tasks shown. Bold: group-best accuracy (ties included). Small colored numbers: change from the corresponding text-only baseline in percentage points (green for gains, red for losses). Janus changes use unrounded accuracies.
}
\label{tab:main-2}
\end{table*}

\newpage

\paragraph{Model Selection.}
We pair every VLM with every image generator in \tabref{tab:models}, and pair Gemini 3.0 Flash~\citep{google2025gemini3flash} with its own Nano-Banana, for 13 modular pipelines in total. We also evaluate BAGEL~\citep{deng2025bagel} and Janus-Pro-7B~\citep{chen2025januspro} as unified models. All models use non-thinking mode unless otherwise specified.

\vspace{-.27cm}

\paragraph{Benchmark Selection.}
We evaluate on nine benchmarks covering six kinds of visual reasoning (\tabref{tab:benchmarks}). We exclude Rover-TG's spatial-context task and ChartQAPro's paragraph-context task.

\vspace{-.27cm}

\paragraph{Evaluation Protocol.}
We score answers by rule where possible. For responses requiring semantic scoring, the original experiments use Gemini-2.5-Flash~\citep{comanici2025gemini}. The Janus-Pro-7B evaluation, matched controls, and extended scaling runs use Gemini-2.5-Flash-Lite. Prompts and inference settings appear in \apprefc{app:implementation} and \apprefc{app:prompts}.

\section{Drawing Is Not Yet Thinking}
\label{sec:results}

\vspace{-.15cm}

\tabref{tab:main-1} and \tabref{tab:main-2} compare every pipeline with the text-only baseline of its VLM on ten tasks: the nine benchmarks, with RoboSpatial split into its compatibility and configuration questions.

\vspace{-.2cm}
\paragraph{Generation usually lowers accuracy.}
Across the 13 modular pipelines, generation lowers accuracy in 106 of the 130 results (82\%); on average, accuracy falls by 3.7 points. In every pipeline, generation lowers accuracy on at least six of the ten tasks, and in 12 of the 13, on at least eight.

\vspace{-.2cm}
\paragraph{Stronger models do not help.}
The pattern holds for all four VLMs and all four generators. Gemini 3.0 Flash, which has the highest text-only accuracy on eight of the ten tasks, loses the most: paired with Nano-Banana, it drops on nine tasks, by 6.9 points on average, and gains on none. Doubling Qwen3-VL from 4B to 8B barely changes the average loss (3.4 vs.\ 3.2 points). Among generators, the strongest and proprietary Nano-Banana lose in 31 of its 40 results.

\vspace{-.2cm}
\paragraph{Losses are largest on visual search.}
On VSTAR, which asks about small objects in high-resolution images, generation lowers accuracy in every pipeline, by 3.8 to 31.3 points and by 21.0 on average. ChartQAPro, MathVista, and ERQA also decline in every pipeline, by 2.4, 5.3, and 4.3 points on average. In the annotated failures (\secrefc{sec:failure_attribution}), generators often lose fine details on ChartQAPro and VSTAR, while the hint often requests the wrong content on MathVista and ERQA.

\vspace{-.2cm}
\paragraph{Gains are confined to one spatial task.}
RoboSpatial-Compatibility is the only task where generation helps consistently: accuracy rises in 12 of the 13 pipelines, by up to 23.8 points, and never falls. A plausible reason is that its questions are hypothetical (would an object fit in a given place?), so drawing the object in place depicts exactly the scene in question. Configuration questions from the same benchmark, which concern the scene as it is, decline in 12 of the 13 pipelines. Elsewhere, the remaining 11 gains are all small, at most 2.2 points.

\vspace{-.2cm}
\paragraph{Both unified models lose accuracy on eight tasks.}
BAGEL gains on RoboSpatial-Compatibility (25.7 points) and MathVista (8.1), and its largest loss is on VSTAR (19.4 points). Across 5,906 paired questions, Janus-Pro-7B loses 9.52 points on RoboSpatial-Compatibility and 9.40 on MathVista. Its largest loss is on ERQA, from 34.50\% to 20.25\%. Its two gains are on VisualPuzzles (1.28 points) and Rover-TG (2.03 points).

\par

\setcounter{topnumber}{1}
\renewcommand{\bottomfraction}{0.5}
\newpage

\section{Why Does Generation Fail to Help?}
\label{sec:reasons}

\begin{figure*}[t]
\centering
\sbox{\figfivepanelabox}{\begin{minipage}{0.49\textwidth}
    \centering
\tablefont\fontsize{8.5}{10.2}\selectfont
\setlength{\tabcolsep}{3.0pt}
\renewcommand{\arraystretch}{1.04}
\ifdefined\fthreelabw\else
  \newlength{\fthreelabw}\newlength{\fthreeaccw}\newlength{\fthreecallw}\newlength{\fthreewd}
\fi
\providecommand{\fthreewidest}[2]{\settowidth{\fthreewd}{#2}\ifdim\fthreewd>#1\setlength{#1}{\fthreewd}\fi}
\setlength{\fthreelabw}{0pt}\setlength{\fthreeaccw}{0pt}\setlength{\fthreecallw}{0pt}
\fthreewidest{\fthreelabw}{\textbf{model / variant}}%
\fthreewidest{\fthreelabw}{\textbf{Qwen3-VL 4B (text)}}%
\fthreewidest{\fthreelabw}{\textbf{Qwen3-VL 8B (text)}}%
\fthreewidest{\fthreelabw}{\quad + \textit{Forced-Gen}}%
\fthreewidest{\fthreelabw}{\quad + \textit{Selective-Gen}}%
\fthreewidest{\fthreeaccw}{\textbf{acc.}}%
\fthreewidest{\fthreeaccw}{35.6}%
\fthreewidest{\fthreecallw}{\textbf{call (\%)}}%
\fthreewidest{\fthreecallw}{100.0}%
\fthreewidest{\fthreecallw}{\textbf{11.5}}%
\makeatletter
\providecommand{\fthreebg}[4]{\rlap{\hspace*{-\tabcolsep}\raisebox{\dimexpr-\dp\@arstrutbox-#4\relax}[0pt][0pt]{%
  \textcolor{#1}{\rule{\dimexpr#2+2\tabcolsep+#3\relax}{\dimexpr\ht\@arstrutbox+\dp\@arstrutbox+#4\relax}}}}}
\makeatother
\providecommand{\fthL}[2]{\fthreebg{#1}{\fthreelabw}{0.8pt}{0.8pt}#2}
\providecommand{\fthA}[2]{\fthreebg{#1}{\fthreeaccw}{0.8pt}{0.8pt}\makebox[\fthreeaccw]{#2}}
\providecommand{\fthC}[2]{\fthreebg{#1}{\fthreecallw}{0.8pt}{0.8pt}\makebox[\fthreecallw]{#2}}
\providecommand{\fthE}[2]{\fthreebg{#1}{\fthreecallw}{0pt}{0.8pt}\makebox[\fthreecallw]{#2}}
\providecommand{\fthLl}[2]{\fthreebg{#1}{\fthreelabw}{0.8pt}{0pt}#2}
\providecommand{\fthAl}[2]{\fthreebg{#1}{\fthreeaccw}{0.8pt}{0pt}\makebox[\fthreeaccw]{#2}}
\providecommand{\fthCl}[2]{\fthreebg{#1}{\fthreecallw}{0.8pt}{0pt}\makebox[\fthreecallw]{#2}}
\providecommand{\fthEl}[2]{\fthreebg{#1}{\fthreecallw}{0pt}{0pt}\makebox[\fthreecallw]{#2}}

\resizebox{\linewidth}{!}{%
\begin{tabular}{w{l}{\fthreelabw}w{c}{\fthreeaccw}w{c}{\fthreecallw}w{c}{\fthreeaccw}w{c}{\fthreecallw}}
\toprule
\textbf{model / variant}
& \multicolumn{2}{c}{\textbf{ChartQAPro}}
& \multicolumn{2}{c}{\textbf{VSTAR}} \\
\cmidrule(lr){2-3} \cmidrule(lr){4-5}
& \textbf{acc.} & \textbf{call (\%)}
& \textbf{acc.} & \textbf{call (\%)} \\
\specialrule{\lightrulewidth}{\aboverulesep}{0pt}
\fthL{basegray}{\textbf{Qwen3-VL 4B (text)}}
& \fthA{basegray}{35.6} & \fthC{basegray}{0.0} & \fthA{basegray}{81.3} & \fthE{basegray}{0.0} \\
\fthL{white}{\quad + \textit{Forced-Gen}}
& \fthA{red!7}{35.5} & \fthC{white}{100.0} & \fthA{red!7}{77.5} & \fthE{white}{100.0} \\
\fthL{white}{\quad + \textit{Selective-Gen}}
& \fthA{white}{35.6} & \fthC{white}{\textbf{9.8}} & \fthA{red!7}{80.6} & \fthE{white}{\textbf{4.0}} \\
\fthL{basegray}{\textbf{Qwen3-VL 8B (text)}}
& \fthA{basegray}{40.8} & \fthC{basegray}{0.0} & \fthA{basegray}{81.6} & \fthE{basegray}{0.0} \\
\fthL{white}{\quad + \textit{Forced-Gen}}
& \fthA{red!7}{38.7} & \fthC{white}{100.0} & \fthA{red!7}{69.6} & \fthE{white}{100.0} \\
\fthLl{white}{\quad + \textit{Selective-Gen}}
& \fthAl{red!7}{40.2} & \fthCl{white}{\textbf{11.5}} & \fthAl{white}{81.6} & \fthEl{white}{\textbf{2.0}} \\
\specialrule{\heavyrulewidth}{0pt}{\belowrulesep}
\end{tabular}%
}
\end{minipage}}
\sbox{\figfivelegendbox}{\includegraphics[width=0.49\textwidth]{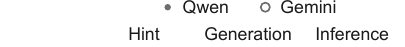}}
\sbox{\figfiveticksbox}{\includegraphics[width=0.49\textwidth]{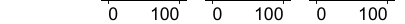}}
\begin{minipage}[t]{0.49\textwidth}
    \centering\offinterlineskip\vspace{0pt}
    \vspace*{\dimexpr\ht\figfivelegendbox+\dp\figfivelegendbox+4pt\relax}
    \raisebox{\dp\figfivepanelabox}{\usebox{\figfivepanelabox}}\par
    \vspace*{\dimexpr\ht\figfiveticksbox+\dp\figfiveticksbox\relax}
    \vspace{0pt}
    \parbox[c][1.6em][c]{\linewidth}{\centering\normalsize\bfseries (a) generation policy ablation}
\end{minipage}
\hfill
\begin{minipage}[t]{0.49\textwidth}
    \centering\offinterlineskip\vspace{0pt}
    \usebox{\figfivelegendbox}\par
    \vspace{4pt}
    \includegraphics[width=\linewidth,height=\dimexpr\ht\figfivepanelabox+\dp\figfivepanelabox\relax]{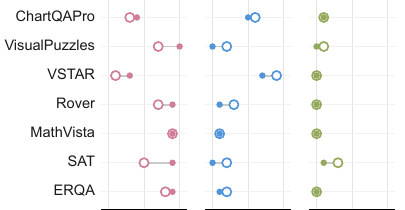}\par
    \usebox{\figfiveticksbox}\par
    \vspace{0pt}
    \parbox[c][1.6em][c]{\linewidth}{\centering\hspace*{44.3pt}\normalsize\bfseries (b) failure attribution}
\end{minipage}
\vspace{-.25cm}
\caption{\textbf{Given the choice, the model rarely generates. Most annotated errors start with the hint.}
\textbf{(a)} Qwen3-VL calls the generator on fewer than 12\% of examples under \emph{Selective-Gen}, while \emph{Forced-Gen} lowers accuracy on both benchmarks, most on VSTAR.
\textbf{(b)} Share (\%) of annotated failures by the step that went wrong, for Qwen (filled) and Gemini (open). Bad hints dominate spatial and embodied tasks. Bad generations dominate ChartQAPro and VSTAR.}
\label{fig:failure_attribution}
\vspace{.3cm}
\end{figure*}

\vspace{-.15cm}
\paragraph{Most annotated failures start with the hint.}
\label{sec:failure_attribution}
We labeled 10--20 failed examples per sampled benchmark by the step that went wrong (\figref{fig:failure_attribution}b). Among these examples, 66\% had a bad hint, 28\% a bad generation, and 6\% an error in the final answer. On ChartQAPro and VSTAR, the generator often lost fine details that the hint identified correctly. On spatial, embodied, puzzle, and STEM tasks, a common error was a generic crop or zoom request that missed the information needed. Per-model results appear in \apprefc{app:failure_attribution_detail}.

\begin{figure}[t]
    \centering
    \includegraphics[width=\linewidth]{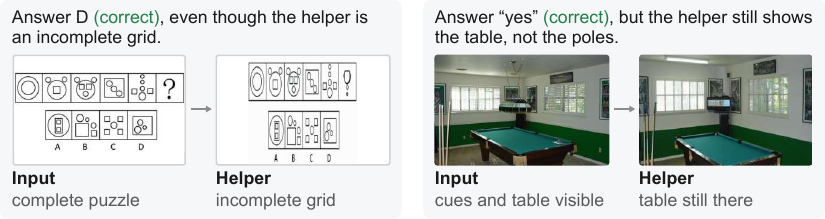}\par\vspace{1pt}
    {\normalsize\bfseries (a) correct answers despite flawed helpers}\par\vspace{8pt}
    \includegraphics[width=\linewidth]{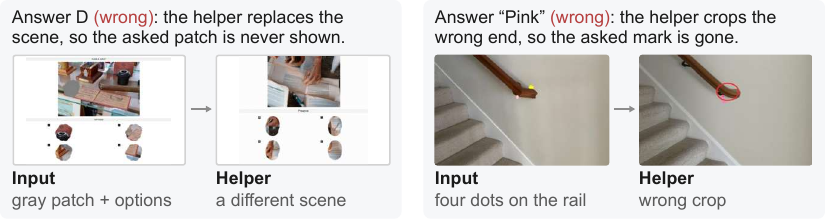}\par\vspace{1pt}
    {\normalsize\bfseries (b) wrong answers: the helper misses the asked region}\par\vspace{8pt}
    \includegraphics[width=\linewidth]{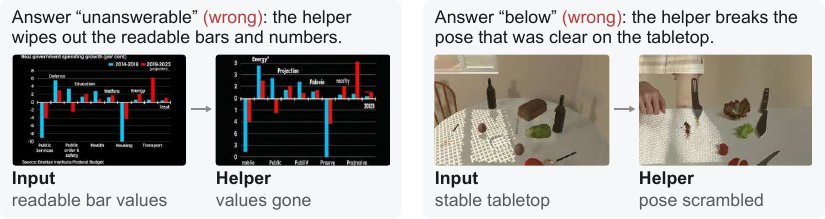}\par\vspace{1pt}
    {\normalsize\bfseries (c) wrong answers: the helper destroys the evidence}\par
    \vspace{-.15cm}
    \caption{\textbf{Generated helper images are rarely what the model needs.}
    Each example shows the model's answer, the input image, and the helper image generated from the model's hint.
    \textbf{(a)} Correct answers occur despite visibly flawed helpers.
    When the answer is wrong, the helper \textbf{(b)} misses the region the question asks about or \textbf{(c)} destroys evidence that was in the input.}
    \label{fig:gen_case_study}
    \vspace{.65cm}
\end{figure}

\vspace{-.13cm}
\paragraph{Examples show the same two failure sources.}
\label{sec:case_study}
\figref{fig:gen_case_study} shows errors in localizing a region, performing an edit, and reproducing a diagram. A model can also answer correctly in the presence of a visibly flawed helper. To measure the helper's contribution, we compare answers to the same question while changing the image in the answer turn (\tabref{tab:matched_helpers}).

\begin{table}[b]
\stepcounter{table}
\centering
\ifdefined\tfivelabw\else\newlength{\tfivelabw}\newlength{\tfivenumw}\newlength{\tfivewd}\fi
\ifdefined\tfivelastw\else\newlength{\tfivelastw}\fi
\makeatletter
\newcommand{\tfivebg}[4]{\rlap{\hspace*{-\tabcolsep}\raisebox{\dimexpr-\dp\@arstrutbox-#4\relax}[0pt][0pt]{%
  \textcolor{#1}{\rule{\dimexpr#2+2\tabcolsep+#3\relax}{\dimexpr\ht\@arstrutbox+\dp\@arstrutbox+#4\relax}}}}}
\makeatother
\newcommand{\tfL}[2]{\tfivebg{#1}{\tfivelabw}{0.8pt}{0.8pt}#2}
\newcommand{\tfN}[2]{\tfivebg{#1}{\tfivenumw}{0.8pt}{0.8pt}\makebox[\tfivenumw]{#2}}
\newcommand{\tfE}[2]{\tfivebg{#1}{\tfivenumw}{0pt}{0.8pt}\makebox[\tfivenumw]{#2}}
\newcommand{\tfLlast}[2]{\tfivebg{#1}{\tfivelabw}{0.8pt}{0pt}#2}
\newcommand{\tfNlast}[2]{\tfivebg{#1}{\tfivenumw}{0.8pt}{0pt}\makebox[\tfivenumw]{#2}}
\newcommand{\tfElast}[2]{\tfivebg{#1}{\tfivenumw}{0pt}{0pt}\makebox[\tfivenumw]{#2}}
\newcommand{\tfM}[2]{\tfivebg{#1}{\tfivelastw}{0pt}{0.8pt}\makebox[\tfivelastw]{#2}}
\newcommand{\tfMlast}[2]{\tfivebg{#1}{\tfivelastw}{0pt}{0pt}\makebox[\tfivelastw]{#2}}
\newcommand{\tfivewidest}[2]{\settowidth{\tfivewd}{#2}\ifdim\tfivewd>#1\setlength{#1}{\tfivewd}\fi}
\sbox{\pairedleft}{\compacttablebody\fontsize{8}{9.6}\selectfont\renewcommand{\arraystretch}{1.12}%
\setlength{\tfivelabw}{0pt}%
\tfivewidest{\tfivelabw}{\textbf{model / variant}}%
\tfivewidest{\tfivelabw}{\textbf{Qwen3-VL 8B (text)}}%
\tfivewidest{\tfivelabw}{+ \textit{controlled prompt}}%
\tfivewidest{\tfivelabw}{+ \textit{oracle visual}}%
\tfivewidest{\tfivelabw}{+ \textit{visual-augmented}}%
\setlength{\tfivenumw}{0pt}%
\tfivewidest{\tfivenumw}{\textbf{jigsaw}}%
\tfivewidest{\tfivenumw}{\textbf{spatial}}%
\tfivewidest{\tfivenumw}{\textbf{maze}}%
\tfivewidest{\tfivenumw}{00.0}%
\setlength{\tfivelastw}{0pt}%
\tfivewidest{\tfivelastw}{\textbf{multi-hop}}%
\tfivewidest{\tfivelastw}{00.0}%
\begin{fittedtabular}{0.49\textwidth}{5}{w{l}{\tfivelabw}*{3}{w{c}{\tfivenumw}}w{c}{\tfivelastw}}
\toprule
\textbf{model / variant}
& \textbf{jigsaw}
& \textbf{spatial}
& \textbf{maze}
& \textbf{multi-hop} \\
\specialrule{\lightrulewidth}{\aboverulesep}{0pt}
\tfL{basegray}{\textbf{Qwen3-VL 8B (text)}} & \tfN{basegray}{47.3} & \tfN{basegray}{42.1} & \tfN{basegray}{44.7} & \tfM{basegray}{65.4} \\
\tfL{white}{+ \textit{controlled prompt}} & \tfN{gainbg}{48.0} & \tfN{gainbg}{44.3} & \tfN{white}{--} & \tfM{white}{--} \\
\tfL{white}{+ \textit{oracle visual}} & \tfN{gainbg}{49.8} & \tfN{white}{--} & \tfN{gainbg}{49.8} & \tfM{gainbg}{67.5} \\
\tfLlast{white}{+ \textit{visual-augmented}} & \tfNlast{dropbg}{46.2} & \tfNlast{dropbg}{37.8} & \tfNlast{dropbg}{40.0} & \tfMlast{dropbg}{63.3} \\
\specialrule{\heavyrulewidth}{0pt}{\belowrulesep}
\end{fittedtabular}%
}
\sbox{\pairedright}{\compacttablebody\fontsize{8}{9.6}\selectfont\renewcommand{\arraystretch}{1.12}%
\setlength{\tfivelabw}{0pt}%
\tfivewidest{\tfivelabw}{\textbf{condition}}%
\tfivewidest{\tfivelabw}{\textbf{text-only}}%
\tfivewidest{\tfivelabw}{+ \textit{visual aug.}}%
\tfivewidest{\tfivelabw}{\textbf{text-only + thinking}}%
\setlength{\tfivenumw}{0pt}%
\tfivewidest{\tfivenumw}{\textbf{MathVista}}%
\tfivewidest{\tfivenumw}{\textbf{ERQA}}%
\tfivewidest{\tfivenumw}{00.0}%
\begin{fittedtabular}{0.49\textwidth}{4}{w{l}{\tfivelabw}*{3}{w{c}{\tfivenumw}}}
\toprule
\textbf{condition}
& \makebox[\tfivenumw][r]{\textbf{ChartQAPro}\hspace{-1pt}}
& \textbf{MathVista}
& \textbf{ERQA} \\
\specialrule{\lightrulewidth}{\aboverulesep}{0pt}
\tfL{basegray}{\textbf{text-only}} & \tfN{basegray}{40.8} & \tfN{basegray}{61.2} & \tfE{basegray}{44.3} \\
\tfL{white}{+ \textit{visual aug.}} & \tfN{dropbg}{38.7} & \tfN{dropbg}{57.3} & \tfE{dropbg}{42.8} \\
\tfL{basegray}{\textbf{text-only + thinking}} & \tfN{basegray}{56.6} & \tfN{basegray}{63.7} & \tfE{basegray}{36.7} \\
\tfLlast{white}{+ \textit{visual aug.}} & \tfNlast{dropbg}{51.2} & \tfNlast{dropbg}{60.8} & \tfElast{dropbg}{28.8} \\
\specialrule{\heavyrulewidth}{0pt}{\belowrulesep}
\end{fittedtabular}%
}

\begin{minipage}[t]{0.49\textwidth}
\centering\vspace{0pt}
\raisebox{\dp\pairedleft}{\usebox{\pairedleft}}\par
\refstepcounter{tablepanel}
(\alph{tablepanel}) controlled and oracle interventions
\label{tab:controlled-analysis}
\end{minipage}\hfill
\begin{minipage}[t]{0.49\textwidth}
\centering\vspace{0pt}
\raisebox{\dp\pairedright}{\usebox{\pairedright}}\par
\refstepcounter{tablepanel}
(\alph{tablepanel}) thinking ablation
\label{tab:thinking_ablation}
\end{minipage}

\addtocounter{table}{-1}
\vspace{-.13cm}
\caption{\textbf{Observed accuracy with fixed hints, reference helpers, and thinking mode.}
Accuracy (\%) of Qwen3-VL 8B.
\textbf{(a)} Nano-Banana generates an image from a fixed instruction for Jigsaw and Spatial (controlled prompt). Qwen receives the reference image directly for Jigsaw, Maze, and Multi-hop (oracle visual).
\textbf{(b)} With FLUX.1-Kontext, generation lowers accuracy with thinking both on and off.
Green/red cells: above/below the gray baseline row of the same block (text-only, or text-only + thinking).}
\label{tab:oracle-thinking}
\end{table}

\vspace{-.13cm}
\paragraph{Fixed hints and reference images.}
\label{sec:upper_bound}
With Qwen3-VL 8B and Nano-Banana, a fixed task-specific hint yields accuracies 0.7 points above the direct-answer baseline on Jigsaw and 2.2 points above it on Spatial (\tabref{tab:controlled-analysis}). Supplying the reference helper directly yields gains of 2.5 points on Jigsaw, 5.1 on Maze, and 2.1 on Multi-hop. The table reports point estimates for each intervention. \apprefc{app:upper_bound_setup} describes both interventions.

\vspace{-.2cm}

\newpage

\paragraph{Changing the helper while holding planning fixed.}
Qwen3-VL 8B receives the same original inputs, hint, turn structure, and final instruction. The answer turn adds either no image, a second copy of the first original, or a FLUX.1-Kontext helper. On VSTAR, accuracy is 60.21\%, 58.64\%, and 49.74\%, respectively. Compared with the hint alone, the generated helper introduces 28 errors and corrects eight, a loss of 10.47 points (paired 95\% CI $[-16.75,-4.71]$). Compared with repeating the original, the loss is 8.90 points ($[-14.66,-2.62]$). The paired intervals span zero on RoboSpatial-Compatibility ($+1.90$ points, $[-5.71,8.57]$) and ChartQAPro ($-0.07$, $[-1.12,0.98]$).

\paragraph{Executing the requested crop and preserving evidence.}
For hints that explicitly request a crop or zoom, Qwen locates the region on the first original image. We crop the original pixels and resize by bicubic interpolation to 1,024 pixels on the long side. On the same 43 VSTAR questions, cropping scores 72.09\% and generation 51.16\%, a 20.93-point advantage for cropping (95\% CI $[4.65,37.21]$). We also instruct the planner to preserve task-relevant objects, text, numbers, and spatial relations. The same preservation hint scores 62.83\% alone and 52.88\% with a generated helper on VSTAR ($-9.95$ points, 95\% CI $[-15.18,-4.19]$). \tabref{tab:matched_helpers} reports all three benchmarks.

\paragraph{How often does the model choose to generate?}
\label{sec:ablation_prompt_control}
By default, the model always generates an image. When Qwen3-VL (4B or 8B) may choose, it calls the generator on 2--12\% of questions and stays within 0.7 points of text-only accuracy. Generating for every question lowers accuracy by up to 12 points on VSTAR (\figref{fig:failure_attribution}a).

\begin{figure}[t]
    \centering
    \includegraphics[width=\linewidth]{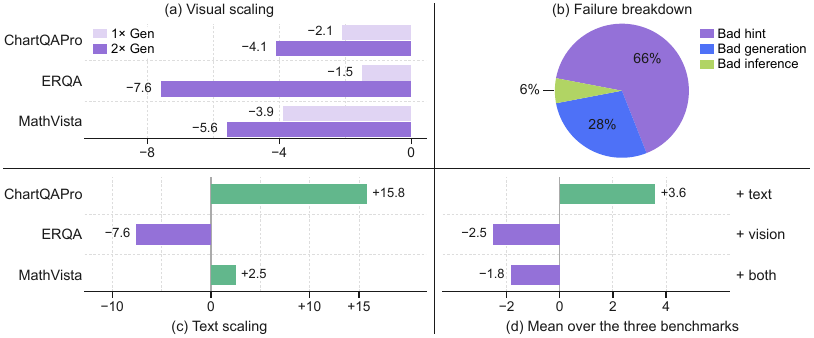}
    \vspace{-.7cm}
    \caption{
    \textbf{Text reasoning and one or two generation calls on three benchmarks.}
    \textbf{(a)} For Qwen3-VL 8B with FLUX.1-Kontext, a second generation call lowers accuracy further on all three benchmarks.
    \textbf{(b)} Share of annotated failures by the step that went wrong.
    \textbf{(c)} Chain-of-thought reasoning raises the same model's accuracy on ChartQAPro and MathVista but lowers it on ERQA.
    \textbf{(d)} Mean change over all three benchmarks from adding text reasoning (as in c), one generation call (1$\times$ Gen in a), or both (\tabref{tab:thinking_ablation}).
    Values in (a), (c), and (d) are changes from the text-only baseline, in percentage points.
    }
    \label{fig:scaling_comparison}
\end{figure}

\vspace{-0.2cm}
\paragraph{The effect of repeated generation depends on the task.}
\label{sec:scaling_language_bias}
For Qwen3-VL 8B with FLUX.1-Kontext, extra text reasoning raises accuracy on ChartQAPro (15.8 points) and MathVista (2.5) and lowers it on ERQA (7.6). In the original one- and two-call comparison, the second generation lowers accuracy on all three (\figref{fig:scaling_comparison}). We reran the zero-, one-, and two-call conditions and extended the sequence to ten edits with a fixed hint (\apprefc{app:extended_visual_scaling}). ChartQAPro and MathVista stay below their no-generation baselines at every evaluated round. ERQA rises from 42.50\% at baseline to 45.50\% after ten edits. Accuracy fluctuates across rounds on all three tasks.

\begingroup
\setlength{\textfloatsep}{8pt plus 1pt minus 2pt}
\vspace{-.2cm}

\paragraph{Thinking does not close the gap.}
\label{sec:thinking}
With thinking mode turned on, generation still lowers accuracy on all three benchmarks (\tabref{tab:thinking_ablation}). The loss grows on ChartQAPro (from 2.1 to 5.4 points) and ERQA (1.5 to 7.9) and shrinks only on MathVista (3.9 to 2.9). Stronger reasoning in text alone does not make the generated images useful. Details are in \apprefc{app:thinking_detail}.

\vspace{-6pt}
\setcounter{topnumber}{2}
\renewcommand{\bottomfraction}{0.3}

\section{Related Work}
\vspace{-4pt}
\paragraph{Multimodal reasoning.}
Following advances in LLMs and large reasoning models~\citep{wei2023chainofthoughtpromptingelicitsreasoning, deepseekai2025deepseekr1incentivizingreasoningcapability, shen2025satorireinforcementlearningchainofactionthought}, VLMs acquire reasoning ability via CoT decomposition~\citep{shao2024visualcotadvancingmultimodal, wang2025multimodal} and reinforcement finetuning~\citep{shen2025vlm, peng2025lmm, sarch2025grounded}. Recent work~\citep{xu2025visual, li2025imagine, wu2025govig} treats images as active reasoning components, extracting task-centric signals such as bounding boxes, depth, and masks~\citep{hu2024visual, shao2024visualcotadvancingmultimodal, liu2025VisualARFT, fu2025refocus} via SFT~\citep{zhang2024provision} or RFT~\citep{zheng2025deepeyes, su2025openthinkimg}. Visual thoughts are typically sourced from tool usage~\citep{fan2026gritteachingmllmsthink, hu2024visual, shao2024visualcotadvancingmultimodal, zheng2025deepeyes, su2025openthinkimg} or code execution~\citep{vinker2024sketchagent, zhang2024provision, liu2025VisualARFT, fu2025refocus}, both limited by tool coverage and the original input. Only a few works~\citep{xu2025visual, li2025imagine, wu2025govig, gu2025thinkmorph} explore image generation as an alternative, and only on narrow domains; whether models can reliably use generated visuals at scale remains open.

\vspace{-4pt}
\paragraph{Unified understanding and generation.}
Unified multimodal models jointly handle understanding and generation in a single architecture~\citep{tong2024metamorph, wu2024vila, ma2025unitok, chen2025blip3}, spanning autoregressive~\citep{dong2023dreamllm, team2024chameleon, chern2024anole, wu2024liquid, wang2024illume, pan2025transfer, xie2025showo2}, diffusion-based~\citep{yang2025mmada, wang2025fudoki, swerdlow2025unified, shi2025muddit}, and hybrid~\citep{zhou2024transfusion, xie2024show, ma2024janusflow, deng2025bagel, liao2025mogao} designs. Building on LLM post-training~\citep{deepseekai2025deepseekr1incentivizingreasoningcapability}, recent efforts apply CoT and RL to unified models for image editing and reasoning generation~\citep{duan2025got, chern2025thinking, jiang2025t2i, xiao2025mindomni}, showing gains when thinking in text before generating in vision. ThinkMorph~\citep{gu2025thinkmorph} extends this to visual-augmented reasoning using specialized data. We hold each reasoner fixed and compare its answers with and without its generated helper. The two unified models gain on different tasks.

\vspace{-6pt}
\section{Conclusion}
\label{sec:conclusion}
\vspace{-6pt}

Generated helpers lower accuracy in most of the modular and unified settings we tested. Even the two unified models gain on different tasks. On VSTAR, holding the hint fixed isolates a loss from the generated image, and cropping the requested region from the original pixels gives better answers. Repeated edits have task-dependent effects. ChartQAPro and MathVista remain below their no-generation baselines through ten edits, while ERQA finishes above baseline. These comparisons can also be applied to video reasoning and planning, where the evidence needed may change from one step to the next.

\par
\endgroup

\bibliographystyle{iclr2027_conference}
\bibliography{main}

\newpage
\section*{Appendix}

\appendix

\noindent This appendix provides additional methodology details, evaluation protocols, and per-benchmark analyses supporting the main paper:
\begingroup

\begin{itemize}
    \item \S\,\hyperref[app:implementation]{A} details the modular VLM--generator pipeline and unified-model inference setup, including decoding hyperparameters and benchmark adjustments.
    \item \S\,\hyperref[app:prompts]{B} lists all prompts used for text-only inference, generation-hint planning, visual-augmented reasoning, and the LLM judge.
    \item \S\,\hyperref[app:additional]{C} reports generator and judge evaluations, inference implementations, failure attribution, thinking-mode results, extended visual scaling, and matched helper-image controls.
    \item \S\,\hyperref[app:limitations]{D} discusses the limitations of our study and adjacent paradigms left to future work.
    \item \S\,\hyperref[app:broader_impact]{E} considers broader societal impacts of generation-augmented multimodal reasoning.
    \item \S\,\hyperref[app:licenses]{F} lists all dataset licenses, source URLs, and terms of use.
\end{itemize}
\endgroup

\begingroup
\hypersetup{linkcolor=black}
\setlength{\parindent}{0pt}
\setlength{\parskip}{1.5pt}
\newcommand{\apptocleaders}{\leaders\hbox to .6em{\hss.\hss}\hfill}
\newcommand{\apptocsec}[3]{%
  \par\addvspace{3pt}%
  \hyperref[#2]{\makebox[1.7em][l]{\textbf{#1}}\textbf{#3}}\nobreak\apptocleaders\nobreak
  \hbox to 1.8em{\hss\textbf{\pageref{#2}}}\par}
\newcommand{\apptocsub}[3]{%
  \par\hspace*{1.7em}\hyperref[#2]{\makebox[2.3em][l]{#1}#3}\nobreak\apptocleaders\nobreak
  \hbox to 1.8em{\hss\pageref{#2}}\par}
\vspace{6pt}
\hrule height .4pt
\vspace{6pt}
{\large\textbf{Contents}}\par
\vspace{2pt}
\apptocsec{A}{app:implementation}{Implementation Details}
\apptocsec{B}{app:prompts}{Evaluation Prompts}
\apptocsec{C}{app:additional}{Additional Experiments}
\apptocsub{C.1}{app:gen_quality}{Generator Capability}
\apptocsub{C.2}{app:judge_sensitivity}{Judge Reliability}
\apptocsub{C.3}{app:kvcache}{Multi-Turn vs.\ KV-Cache Inference}
\apptocsub{C.4}{app:failure_attribution_detail}{Detailed Failure Attribution}
\apptocsub{C.5}{app:case_study_detail}{Detailed Qualitative Case Studies}
\apptocsub{C.6}{app:thinking_detail}{Detailed Thinking-Mode Results}
\apptocsub{C.7}{app:extended_visual_scaling}{Extended Visual Scaling}
\apptocsub{C.8}{app:matched_controls}{Matched Helper-Image Controls}
\apptocsec{D}{app:limitations}{Limitations}
\apptocsec{E}{app:broader_impact}{Broader Impact}
\apptocsec{F}{app:licenses}{Dataset Licenses and Asset Credits}
\vspace{6pt}
\hrule height .4pt
\endgroup

\section{Implementation Details}
\label{app:implementation}

This section summarizes evaluation settings for both modular VLM--generator pipelines and unified multimodal models.

\paragraph{Text-only and unified model inference.}
For all VLM evaluations, we fix the random seed for reproducibility. Unless otherwise specified, decoding uses a temperature of $0.1$, top-$p$ sampling with $p=1.0$, and a maximum token budget of $8192$, which is sufficient to accommodate the longer reasoning chains required by benchmarks such as MathVista and VisualPuzzles without truncation.

\paragraph{Reasoning with Imagination Framework.}
In the modular VLM--generator setting, inference proceeds in two stages. In \textbf{Turn 1}, the VLM produces a generation hint describing the desired auxiliary visual content. We use a temperature of $0.1$, top-$p=1.0$, and a token budget of $2048$ for this stage. The hint is then passed to an external image generator to produce a helper image.

For Stable Diffusion 3.5, we use 28 denoising steps, guidance scale $5.0$, and strength $0.6$ for image-to-image generation. For FLUX.1-Kontext, we use 28 denoising steps, guidance scale $2.5$, and CFG scale $1.0$. In \textbf{Turn 2}, the VLM receives both the original inputs and the generated helper image and produces the final answer using the same decoding settings as in text-only inference. Detailed hyper-parameter settings are provided in \tabref{tab:impl_hparams}.

BAGEL uses its official inference settings. For Janus-Pro-7B, we prefix the generation head with image and instruction embeddings, sample one $384\times384$ helper (576 image tokens, guidance 5, temperature 1), and decode text greedily. The direct and generation conditions use the same questions.

\begin{table}[htbp]
\centering
\begingroup
\tablebody
\begin{fittedtabular}{0.82\linewidth}{3}{llr}
\toprule
\textbf{component} & \textbf{parameter} & \textbf{value} \\
\midrule
\multirow{3}{*}{\textbf{text inference (VLM / unified)}} 
& temperature & 0.1 \\
& top-$p$ & 1.0 \\
& max tokens & 8192 \\
\midrule
\multirow{3}{*}{\textbf{turn 1 hint generation}}
& temperature & 0.1 \\
& top-$p$ & 1.0 \\
& max tokens & 2048 \\
\midrule
\multirow{3}{*}{\textbf{Stable Diffusion 3.5}}
& denoising steps & 28 \\
& guidance scale & 5.0 \\
& strength & 0.6 \\
\midrule
\multirow{3}{*}{\textbf{FLUX.1-Kontext}}
& denoising steps & 28 \\
& guidance scale & 2.5 \\
& CFG scale & 1.0 \\
\bottomrule
\end{fittedtabular}
\endgroup

\caption{Default inference hyperparameters for the modular pipelines. Settings for Janus-Pro-7B and the extended scaling runs are given in the text.}
\label{tab:impl_hparams}
\end{table}

\paragraph{Benchmark adjustments.}
To reduce ambiguity in evaluation and focus on reasoning-centric tasks, we use the Compatibility and Configuration splits from RoboSpatial and exclude the embodied-interaction tasks from Rover-Bench, as these settings require environment dynamics or action feedback that are not compatible with our static generation framework.

\paragraph{Oracle-Conditions Setup (\secrefc{sec:upper_bound}).}
\label{app:upper_bound_setup}
The controlled tasks draw on visual perception, spatial relations, maze navigation, and counting benchmarks~\citep{ray2024sat, liang2025rover, li2025zebra, hao2025can, shi2025realunifyunifiedmodelstruly}. Reference information is provided by the benchmark~\citep{shi2025realunifyunifiedmodelstruly, liang2025rover, zou2025uni} or constructed for the task. For Jigsaw and Spatial, we replace the model-generated hint with a fixed task-specific instruction and use Nano-Banana to generate the helper. For Jigsaw, Maze, and Multi-hop, we bypass generation and supply the reference helper image directly to Qwen3-VL 8B (\tabref{tab:controlled-analysis}).

\paragraph{Test-Time Scaling Setup (\secrefc{sec:scaling_language_bias}).}
\label{app:scaling_setup}
\emph{Text scaling} adds a chain-of-thought prompt at inference time. \emph{Visual scaling} varies the number of generator calls. \figref{fig:scaling_comparison} reports the original one- and two-call results. \apprefc{app:extended_visual_scaling} gives the repeated-edit protocol, token limits, and per-call timing for the extended run.

\section{Evaluation Prompts}
\label{app:prompts}
\paragraph{Text-only Direct prompting.}
We do not add additional instructions unless otherwise specified. A typical prompt follows the format:
\begin{tcolorbox}[promptbox]
You are given a visual reasoning problem based on the given image(s).

Question: \{Question\}  

options: \{Options\}(if exists)

Return only the final answer and Wrap your final answer in $boxed\{\}$.
\end{tcolorbox}

\paragraph{Text-only Chain-of-Thought prompting.}
For text-side scaling in \secrefc{sec:reasons}, we use the following chain-of-thought prompt:
\begin{tcolorbox}[promptbox]
You are given a visual reasoning problem based on the given image(s).

Question: \{Question\}  

options: \{Options\}(if exists)

Think step by step, carefully reasoning through the problem before giving your answer.

Wrap your final answer in $boxed\{\}$.
\end{tcolorbox}

\paragraph{Visual-augmented reasoning: Turn 1 (generation hint) for Flux.1 and SD3.5.}
The VLM first produces a short instruction to guide the external image generator. Since Flux.1 and SD3.5 are built for dealing with only one input image, we specifically prompt the edit instruction on the first input image:
\begin{tcolorbox}[promptbox]
You will be given a question and one or more input images. Do NOT answer the question yet.

An external tool can generate ONE helper image, but it can ONLY take the FIRST image as the init image.

So your instruction must describe how to edit the FIRST image into a helper image.
If there are multiple images, you may use them as reference, but the helper must be generated from the FIRST image.

Write a SHORT instruction (1-2 sentences) for generating the helper image.

Wrap your instruction in <instruction> tags:<instruction>Your image generation instruction here.</instruction>

Do NOT answer the question yet.
\end{tcolorbox}

\paragraph{Visual-augmented reasoning: Turn 1 (generation hint) for Qwen-Image-Edit and Nano-banana.}
Qwen-Image-Edit-2511 and Nano-banana can take in multiple images for input:
\begin{tcolorbox}[promptbox]
You will be given a question and one or more input images. Do NOT answer the question yet.

You can call an external tool that will generate ONE helper image.

The tool can see ALL the original images as context, so your instruction may reference any of them.

Write a SHORT instruction (1-2 sentences) for generating the helper image.

Wrap your instruction in <instruction> tags: <instruction>Your image generation instruction here.</instruction>

Do NOT answer the question yet.
\end{tcolorbox}

\paragraph{Visual-augmented reasoning: Turn 1 (generation hint) CoT prompt.}
For the additional experiment in \secrefc{sec:thinking}, we add a thinking prompt to the hint-generation turn. The final-answer turn uses the text-only CoT prompt.
\begin{tcolorbox}[promptbox]
Before writing your instruction, think step by step about what visual information is missing or hard to extract from the original image(s), and how a generated helper image could make answering the question easier.
\end{tcolorbox}

\paragraph{Visual-augmented reasoning: Turn 2 (final answer).}
After the helper image is generated, the VLM receives both original and generated images:
\begin{tcolorbox}[promptbox]
You will see (A) the ORIGINAL input image(s) and (B) an OPTIONAL GENERATED helper image.

Answer using evidence from all provided image(s).
\end{tcolorbox}

\paragraph{LLM-as-Judge evaluation prompt.}
For benchmarks without rule-based evaluation, we use the following prompt to assess answer correctness:
\begin{tcolorbox}[promptbox]
You are a strict evaluation judge for QA tasks.

Goal: decide if the model prediction should be counted as correct for the given sample.

Rules:

- If options are provided, the prediction is correct iff it clearly selects the same option as the ground-truth answer.

- Accept either option letter (A, B, C...) or the option text, as long as it maps unambiguously.

- Be robust to minor formatting differences (case, punctuation, whitespace).

- For open-ended answers without options, mark correct only if the prediction is semantically equivalent to the ground truth.

- If prediction is ambiguous, incomplete, or picks multiple conflicting answers, mark incorrect.

- Do NOT be lenient: if you are unsure, mark incorrect.

Return ONLY JSON with keys:

  correct: boolean

  matched\_option\_index: integer or null  (0-based if options exist)

  normalized\_pred: string

  normalized\_answer: string

  rationale: string  (one short sentence)

\end{tcolorbox}

\section{Additional Experiments}
\label{app:additional}

\subsection{Generator Capability}
\label{app:gen_quality}
\label{sec:ablation_generator_difficulty}
We further investigate whether the limited gains of visual generation stem from the generator's capability rather than the reasoning model. Concretely, we perform a focused case study that stratifies tasks by difficulty and contrasts scenarios where image generation is expected to be sufficient (simple perceptual edits) against those requiring complex manipulations or structured visual outputs. Unless otherwise specified, experiments in this subsection use Qwen3-VL 8B~\citep{bai2025qwen3vl} as the reasoning model and Nano-Banana~\citep{google2025gemini25flashimage} as the image generator. We chose two tasks, a perception-heavy jigsaw task and a structured math-geometry task, and control the generation hint to be minimal short prompts without any reasoning involved.
\begin{table}[t]
\centering
\begingroup
\tablebody
\begin{fittedtabular}{0.82\linewidth}{5}{lrrrr}
\toprule
\textbf{setting}
& \multicolumn{2}{c}{\textbf{jigsaw}}
& \multicolumn{2}{c}{\textbf{math-geo}} \\
\cmidrule(lr){2-3} \cmidrule(lr){4-5}
& \textbf{easy} & \textbf{hard}
& \textbf{easy} & \textbf{hard} \\
\midrule

\rowcolor{basegray}
\textbf{text-only}
& 76.8 & 34.2
& 55.2 & 23.1 \\

\quad + \textit{generation}
& 77.1 {\scriptsize(+0.3)}
& \drop{31.0 {\scriptsize(-3.2)}}
& 54.9 {\scriptsize(-0.3)}
& \drop{19.8 {\scriptsize(-3.3)}} \\

\bottomrule
\end{fittedtabular}%
\endgroup

\caption{
Generation degrades performance primarily on hard, structured instances.
On easy examples, generation leaves accuracy nearly unchanged, whereas
performance drops by over 3 points on the hard subsets of both the jigsaw and
math-geo tasks. Accuracy (\%) is reported for Qwen3-VL 8B with Nano-Banana;
parentheses show changes relative to text-only inference.
}
\label{tab:gen_difficulty}

\end{table}

\tabref{tab:gen_difficulty} shows difficulty-stratified results. Generation provides marginal benefit on easy perceptual instances (e.g., jigsaw easy: 76.8$\to$77.1) but this benefit diminishes or reverses on hard and structured instances (math-geo hard: 23.1$\to$19.8), consistent with the failure modes identified in \secrefc{sec:failure_attribution}.

\subsection{Judge Reliability}
\label{app:judge_sensitivity}

\begin{table}[t]
\centering
\begingroup
\tablebody
\begin{fittedtabular}{0.82\linewidth}{4}{lrrr}
\toprule
\textbf{judge}
& \textbf{ChartQAPro}
& \textbf{VSTAR}
& \textbf{audit set} \\
\midrule

\rowcolor{basegray}
\textbf{rule-based}
& 40.8 & 81.6 & -- \\

Gemini-2.5-Flash
& 41.2 {\scriptsize(+0.4)}
& 81.6 {\scriptsize(+0.0)}
& 37.0 {\scriptsize(-2.0)} \\

\rowcolor{basegray}
\textbf{human}
& -- & -- & 39.0 \\

\bottomrule
\end{fittedtabular}%
\endgroup

\caption{
LLM-based evaluation closely agrees with rule-based and human scoring.
Gemini-2.5-Flash differs from rule-based accuracy by at most 0.4 points
on ChartQAPro and VSTAR, and by 2.0 points from human annotation on a
100-instance audit set sampled across benchmarks.
Results are based on Qwen3-VL 8B text-only predictions.
The cross-run standard deviation of the LLM judge is 0.97 points.
}
\label{tab:judge_reliability}

\end{table}

We use Gemini-2.5-Flash as an LLM judge for benchmarks where rule-based matching is not applicable. To assess reliability, we compare it against (1) rule-based evaluation on benchmarks with deterministic answer formats, and (2) human annotations on a randomly sampled set of 100 instances drawn from all datasets. All results in \tabref{tab:judge_reliability} use Qwen3-VL 8B text inference.

\paragraph{Consistency with rule-based evaluation.}
The two evaluators yield closely matching aggregate accuracies, with identical scores on VSTAR and a difference of only 0.4 percentage points on ChartQAPro.

\paragraph{Consistency with human annotation.}
On the 100-instance sample set, Gemini-2.5-Flash scores 37.0 versus human evaluation at 39.0, a 2-point difference well within the margin of inter-annotator variation. We additionally rerun Gemini-2.5-Flash three times on ChartQAPro and VSTAR, observing a standard deviation of 0.97 percentage points across runs. Together, these checks confirm that our evaluation judge is robust and that findings are not an artifact of judge design.

\subsection{Multi-Turn vs.\ KV-Cache Inference}
\label{app:kvcache}

Once the helper image has been produced, we compare two implementations for feeding it back to the VLM. \emph{Multi-turn} inference adds it in a new conversational turn and re-encodes the full context, including the question, original images, hint, and helper image. \emph{KV-cache} inference instead reuses the cached prefix and processes the appended image tokens. We compare these implementations on VSTAR, ERQA, and SAT using Qwen3-VL 8B paired with Qwen-Image-Edit.

\begin{table}[t]
\centering
\begingroup
\tablebody
\begin{fittedtabular}{0.82\linewidth}{4}{lrrr}
\toprule
\textbf{strategy}
& \textbf{VSTAR}
& \textbf{ERQA}
& \textbf{SAT} \\
\midrule

\rowcolor{basegray}
\textbf{text-only}
& 81.6 & 44.3 & 70.1 \\

\quad + \textit{gen (KV-cache)}
& \drop{32.4}
& \drop{18.7}
& \drop{45.8} \\

\quad + \textit{gen (multi-turn)}
& \drop{50.8}
& \drop{38.5}
& \drop{61.6} \\

\bottomrule
\end{fittedtabular}%
\endgroup

\caption{
Multi-turn inference substantially outperforms KV-cache reuse,
but accuracy stays below the text-only baseline.
Re-encoding the generated image in a new turn improves accuracy over
KV-cache reuse by 15--20 points across all three benchmarks.
Accuracy (\%) is reported for Qwen3-VL 8B with Qwen-Image-Edit.
}
\label{tab:kvcache_vs_multiturn}

\end{table}

Across all three benchmarks, the multi-turn implementation outperforms the KV-cache implementation: 50.8 vs.\ 32.4 on VSTAR (+18.4), 38.5 vs.\ 18.7 on ERQA (+19.8), and 61.6 vs.\ 45.8 on SAT (+15.8). We therefore use multi-turn inference for all main experiments. These results compare the two evaluated implementations; they do not establish an inherent limitation of KV caching. Under causal attention, recomputing an unchanged prefix does not let earlier tokens attend to a later helper image. The source of the observed gap remains unresolved.

Even with multi-turn inference, accuracy remains below the text-only baseline (81.6 / 44.3 / 70.1). Thus, the main finding holds under the better-performing implementation: generated helper images still reduce accuracy on these three benchmarks.

\subsection{Detailed Failure Attribution}
\label{app:failure_attribution_detail}

In the annotated sample, \textbf{76\%} of Qwen3-VL failures and \textbf{57\%} of Gemini failures stem from bad hints, while \textbf{21\%} (Qwen) and \textbf{36\%} (Gemini) come from bad generation; pure inference failures are negligible. Per-task ranges sharpen the dichotomy reported in \secrefc{sec:failure_attribution}: chart and visual-search tasks (ChartQAPro, VSTAR) reach 50--90\% bad generation, while spatial, embodied, puzzle, and STEM tasks (ERQA, MathVista, SAT, VisualPuzzles, Rover) reach 70--100\% bad hints. Stylistically, Gemini more frequently produces tool-like hints (e.g., Python-style editing instructions) while Qwen3-VL produces shorter, under-specified ones; despite this stylistic difference, both models exhibit the same task-dependent split.

\subsection{Detailed Qualitative Case Studies}
\label{app:case_study_detail}

In \figref{fig:gen_case_study}, a correct answer sometimes accompanies an inaccurate helper. The examples identify visible errors in the helper. The paired comparisons in \apprefc{app:matched_controls} measure how often a different helper changes the outcome for the same question. Among failed cases, some hints identify the wrong region. Other hints identify the right evidence, but the generator alters it during a precise edit or while reconstructing a diagram.

\subsection{Detailed Thinking-Mode Results}
\label{app:thinking_detail}

For text-only inputs, thinking yields ChartQAPro $+$15.8 (40.8$\to$56.6) and MathVista $+$2.5 (61.2$\to$63.7) but hurts ERQA ($-$7.6, 44.3$\to$36.7). Adding visual augmentation without thinking already degrades all three (FLUX.1: ChartQAPro $-$2.1, MathVista $-$3.9, ERQA $-$1.5, from 44.3 to 42.8), and enabling thinking does not recover the gap and sometimes widens it ($-$5.4, $-$2.9, $-$7.9 respectively). On harder reasoning tasks, longer reasoning over the generated image may even amplify the noise the synthetic visual injects.

\subsection{Extended Visual Scaling}
\label{app:extended_visual_scaling}

\tabref{tab:extended_visual_scaling} reports a rerun of the no-generation baseline and the one- and two-call conditions with Qwen3-VL 8B and FLUX.1-Kontext, extended to ten calls. We generate one hint per question and keep it fixed. The first call edits the first original image. Each subsequent call edits the preceding helper. At each checkpoint the VLM answers the question afresh using the original images and only the latest helper. At zero calls it answers directly from the originals.

For $n>0$, the budget at round $n$ is one planning call (up to 256 tokens), $n$ image-generation calls, and one answer call (up to 1,024 tokens) for that checkpoint. Each FLUX call uses 24 denoising steps and takes approximately 47--49 seconds on an NVIDIA L40, excluding planning and answering.

\begin{table}[!htbp]
\centering
\begingroup
\tablebody
\begin{fittedtabular}{\linewidth}{9}{lrrrrrrrr}
\toprule
\textbf{Generation calls} & \textbf{0} & \textbf{1} & \textbf{2} & \textbf{3} & \textbf{4} & \textbf{5} & \textbf{8} & \textbf{10} \\
\midrule
ChartQAPro & 39.16 & 37.61 & 37.89 & 37.82 & 38.11 & 38.39 & 37.82 & 37.40 \\
ERQA & 42.50 & 42.50 & 44.00 & 43.25 & 43.75 & 42.75 & 44.00 & 45.50 \\
MathVista & 67.70 & 65.90 & 66.60 & 65.00 & 66.30 & 66.40 & 66.90 & 66.30 \\
\bottomrule
\end{fittedtabular}
\endgroup
\caption{\textbf{Accuracy (\%) across repeated edits.} Each checkpoint uses the same 1,425 ChartQAPro, 400 ERQA, and 1,000 MathVista questions. At zero calls the VLM answers directly from the originals.}
\label{tab:extended_visual_scaling}
\end{table}

\subsection{Matched Helper-Image Controls}
\label{app:matched_controls}

\begin{table}[!htbp]
\centering
\begingroup
\tablebody
\begin{fittedtabular}{\linewidth}{4}{lrrr}
\toprule
\textbf{Condition or comparison} & \textbf{VSTAR} & \textbf{RS-Compat.} & \textbf{ChartQAPro} \\
\midrule
\multicolumn{4}{l}{\textbf{Same hint and answer turn}} \\
Questions & 191 & 105 & 1,425 \\
Hint only & 60.21 & 59.05 & 39.86 \\
Repeated original & 58.64 & 52.38 & 39.79 \\
Generated helper & 49.74 & 60.95 & 39.79 \\
\midrule
\multicolumn{4}{l}{\textbf{Explicit crop or zoom requests}} \\
Eligible questions & 43 & 21 & 2 \\
Hint only & 69.77 & 66.67 & 50.00 \\
Crop of original pixels & 72.09 & 47.62 & 50.00 \\
Generated helper & 51.16 & 52.38 & 50.00 \\
\midrule
\multicolumn{4}{l}{\textbf{Evidence-preserving hint, all questions}} \\
Hint only & 62.83 & 53.33 & 37.33 \\
Generated helper & 52.88 & 55.24 & 38.60 \\
\midrule
\multicolumn{4}{l}{\textbf{Errors introduced / corrected by generation}} \\
Compared with hint only & 28 / 8 & 6 / 8 & 28 / 27 \\
Compared with repeated original & 26 / 9 & 3 / 12 & 22 / 22 \\
Compared with crop (eligible only) & 12 / 3 & 1 / 2 & 0 / 0 \\
\bottomrule
\end{fittedtabular}
\endgroup
\caption{\textbf{Matched helper-image controls for Qwen3-VL 8B with FLUX.1-Kontext.} Accuracies are percentages. The last block counts correct-to-incorrect and incorrect-to-correct transitions. RS-Compat.\ denotes RoboSpatial-Compatibility.}
\label{tab:matched_helpers}
\end{table}

Within each comparison we reuse the original inputs, the same planning hint, and the same two-turn conversation. The final instruction asks the VLM to use the originals and consider any helper provided. We decode answers greedily and estimate 95\% confidence intervals by resampling paired questions 2,000 times. The direct-answer baseline in the main tables answers before planning. The hint-only control answers after planning.

For an explicit crop or zoom request, Qwen3-VL 8B predicts the bounding box from the fixed hint and the first image without access to the reference answer. For the preservation condition, we ask the planner to retain the objects, attributes, text, numbers, labels, scales, and spatial relationships needed to answer the question. We compare the resulting hint alone with the same hint and its generated helper.

\section{Limitations}
\label{app:limitations}

Our study evaluates static image generators and omits two adjacent paradigms that may unlock cases where stills fall short: \textbf{video generators} and \textbf{world-model / simulator generators}, which can render dynamics, action consequences, and counterfactual rollouts that single-frame edits cannot. Whether these temporally grounded generators sidestep the hint and image errors observed here is an open question we leave to future work. Our evaluation is also confined to \textbf{question-answering benchmarks}; visual generation may play a different and possibly more load-bearing role in non-QA settings such as long-horizon planning, agentic execution, dialog, or open-ended generation, where intermediate visual artifacts are themselves part of the output rather than an aid to a single answer. Extending the framework to instruction-tuned reasoning-specialist VLMs and to longer multi-turn text--image--text--image trajectories is another natural direction.

\section{Broader Impact}
\label{app:broader_impact}
This work studies the integration of visual generation into multimodal reasoning systems and finds that generated visual evidence does not consistently improve reasoning and may even degrade performance in structured tasks. These insights can inform the design of more reliable systems and guide when generation should be used.

At the same time, incorporating generated images introduces risks. Generated content may contain hallucinated or misleading details, which could be interpreted as valid evidence, especially in high-stakes settings such as healthcare or autonomous systems. There is also potential for misuse, such as generating visually plausible but incorrect evidence.

We encourage future work on improving the reliability of visual generation, incorporating verification mechanisms, and explicitly modeling uncertainty in multimodal reasoning systems.

\section{Dataset Licenses and Asset Credits}
\label{app:licenses}

We use the following publicly released benchmarks under their respective licenses. All datasets are used only for academic research and we comply with their terms.
\begin{itemize}
    \item \textbf{ChartQAPro}~\citep{masry2025chartqapro} -- MIT License (\url{https://github.com/vis-nlp/ChartQAPro}).
    \item \textbf{V*Bench / VSTAR}~\citep{wu2024v} -- MIT License (\url{https://github.com/penghao-wu/vstar}).
    \item \textbf{VisuLogic}~\citep{xu2025visulogic} -- Apache 2.0 (\url{https://github.com/VisuLogic-Benchmark/VisuLogic-Eval}).
    \item \textbf{VisualPuzzles}~\citep{song2025visualpuzzles} -- MIT License (\url{https://github.com/neulab/VisualPuzzles}).
    \item \textbf{SAT}~\citep{ray2024sat} -- MIT License (\url{https://huggingface.co/datasets/array/SAT}).
    \item \textbf{RoboSpatial}~\citep{song2025robospatial} -- NVIDIA Source Code License, non-commercial research use only (\url{https://github.com/NVlabs/RoboSpatial/blob/main/LICENSE}).
    \item \textbf{ERQA}~\citep{team2025gemini} -- CC-BY 4.0 (\url{https://github.com/embodiedreasoning/ERQA}).
    \item \textbf{MathVista}~\citep{lu2024mathvista} -- CC-BY-SA 4.0 (\url{https://huggingface.co/datasets/AI4Math/MathVista}).
    \item \textbf{Rover-Bench}~\citep{liang2025rover} -- MIT License (\url{https://github.com/cheryyunl/ROVER}).
    \item \textbf{Zebra-CoT}~\citep{li2025zebra} -- CC-BY-NC 4.0, non-commercial research use only (\url{https://huggingface.co/datasets/multimodal-reasoning-lab/Zebra-CoT}).
    \item \textbf{EMMA-mini}~\citep{hao2025can} -- dataset license not specified in the public dataset card (\url{https://huggingface.co/datasets/luckychao/EMMA-mini}).
    \item \textbf{RealUnify}~\citep{shi2025realunifyunifiedmodelstruly} -- Apache 2.0 (code) with academic-research-only restriction on dataset usage (\url{https://github.com/FrankYang-17/RealUnify}).
\end{itemize}

We also use the following models under their official releases: Qwen3-VL~\citep{bai2025qwen3vl}, GLM-4.6V~\citep{hong2025glm}, Gemini 3.0 Flash~\citep{google2025gemini3flash} and Nano-Banana~\citep{google2025gemini25flashimage}, Stable Diffusion 3.5~\citep{esser2024scaling}, FLUX.1-Kontext~\citep{flux2024}, Qwen-Image-Edit~\citep{wu2025qwen}, BAGEL~\citep{deng2025bagel}, and Janus-Pro-7B~\citep{chen2025januspro}.

\end{document}